\documentclass[sigconf, nonacm]{acmart}
\usepackage{multirow}

\begin{document}

\title{RMS@CC-MMD 2026: Multimodal Misogyny Detection via Geometric Interaction and Multi-View Consensus}



\author{Md. Ajwad Hossain}
\affiliation{%
  \department{Department of Electronics and Telecommunication Engineering}
  \institution{Chittagong University of Engineering \& Technology (CUET)}
  \city{Chattogram}
  \country{Bangladesh}
}
\email{md.ajwadhossain@gmail.com}

\begin{abstract}
The proliferation of internet memes has introduced new complexities
to automated content moderation, particularly in detecting misogyny.
Memes often rely on a semantic clash between visual and textual
modalities, where hateful intent is implicit and culturally grounded.
This paper presents \textbf{GeoMVC} (Geometric Interaction and
Multi-View Consensus), developed for the CC-MMD Grand
Challenge at ICMI 2026. To address the limitations of
static feature concatenation, a Geometric
Interaction Layer is proposed that models cross-modal alignment via Hadamard
products and cosine similarity between frozen visual and textual
embeddings. We further mitigate distribution shifts caused by noisy
OCR and code-mixed transliteration through a Multi-View
Consensus strategy, aggregating predictions across raw,
length-filtered, and English-translated text views. The system
achieved \textbf{Rank 2 in the Malayalam partition} (Macro F1:
0.892) and \textbf{Rank 3 in the Chinese partition} (Macro F1:
0.895) on Task A, while securing \textbf{Rank 5 in the Tamil partition} 
(Macro F1: 0.521). A detailed error analysis on the development
partition highlights open challenges in modeling localized
transliteration and code-mixed sarcasm across Dravidian and
Chinese cultural contexts.
\end{abstract}



\maketitle



\section{Introduction}

Online misogyny has found a new vehicle in internet memes, 
which weaponize the combination of image and text to deliver 
hateful or demeaning messages that neither modality would 
convey alone. Reliably detecting such content demands models 
capable of reasoning about cross-modal interactions rather 
than surface-level features, a challenge that becomes 
considerably harder in low-resource and code-mixed linguistic 
settings. Benchmark datasets constructed around this problem 
demonstrate that unimodal classifiers are trivially defeated 
by ``benign confounders''---carefully designed variants that 
flip a meme's label by altering only one modality---making 
the detection of \emph{semantic clash} between text and image 
the central challenge of the field~\cite{kiela2020hateful}.

We present \textbf{GeoMVC} (Geometric Interaction and Multi-View
Consensus), submitted to Task~A of the CC-MMD Grand Challenge
at ICMI~2026. GeoMVC pairs frozen vision--language encoders with
a Geometric Interaction Layer and a Multi-View Consensus mechanism
that aggregates predictions across three textual views at inference
time. The system achieves competitive results on the official
leaderboard, with stronger performance on the Malayalam and Chinese
partitions and notable limitations on Tamil under extreme code-mixing.

Our contributions are as follows:
\vspace{-0.5mm}
\begin{enumerate}
    \item A Geometric Interaction Layer combining 
    Hadamard product and cosine similarity over frozen CLIP 
    and mCLIP embeddings to model element-wise cross-modal 
    correlations.
    \item A Multi-View Consensus scheme that applies 
    majority voting over raw, length-filtered, and 
    machine-translated text views to mitigate OCR and 
    transliteration noise at inference time.
    \item An empirical cross-partition analysis revealing 
    where each mechanism succeeds and fails across Malayalam, 
    Chinese, and Tamil, with particular focus on the 
    compounding effects of code-mixing and label imbalance.
\end{enumerate}
\vspace{-1.00mm}
The source code for GeoMVC is publicly available on \href{https://github.com/Ajwad07/geomvc-ccmmd-icmi2026}{GitHub}.

\section{Related Work}

\subsection{Multimodal Hate Speech and Semantic Clash}

The Hateful Memes benchmark established a foundational 
insight: reliably identifying hate in memes requires 
genuine cross-modal reasoning, since benign confounders 
ensure that neither the image nor the text channel alone 
determines the label~\cite{kiela2020hateful}. The 
offensiveness emerges from the \emph{interaction}---a 
caption that appears neutral becomes derogatory when placed 
alongside a particular image, operationalizing the concept 
of semantic clash~\cite{zhong2020classification}. Follow-on 
systems tackled this through multimodal transformers and 
prediction ensembles, narrowing but not closing the gap 
to human-level performance~\cite{lippe2020multimodal, 
velioglu2020detecting}.

\subsection{Vision--Language Models in a Shared Space}

Contrastive pretraining on web-scale image--caption pairs 
produces visual representations that transfer broadly across 
tasks without task-specific supervision~\cite{radford2021learning}. 
Multilingual extensions of this paradigm distill or retrain 
the text encoder to accommodate non-English input while 
keeping the visual space fixed, enabling cross-lingual 
vision--language retrieval~\cite{carlsson2022cross, 
chen2023mclip}. Low-resource settings benefit further from 
caption augmentation strategies that expand training 
coverage through translation and synthetic 
captioning~\cite{santos2023capivara}, though performance 
remains tied to the quality and availability of parallel 
data.

\subsection{Multimodal Fusion and Geometric Interactions}

Capturing subtle cross-modal dependencies calls for fusion 
strategies that go beyond simple concatenation. Bilinear 
methods such as MUTAN decompose the interaction between 
question and visual features via Tucker factorization, 
yielding expressive yet parameter-efficient 
representations~\cite{Ben-Younes2017MUTAN}. Subsequent 
work demonstrated that element-wise multiplicative 
operations and low-rank decompositions generalize this 
capacity to relational reasoning 
tasks~\cite{Ben-Younes2019BLOCK:, Cadène2019MUREL:}, 
establishing Hadamard-type products as a principled tool 
for modeling fine-grained feature interactions across 
modalities.

\subsection{Code-Mixing and Multilingual VLM Limitations}

Multilingual vision--language models typically assume 
well-formed, sentence-level input and rely on machine 
translation for cross-lingual transfer~\cite{carlsson2022cross, 
chen2023mclip}. Both assumptions break down under 
code-mixing, where users interleave scripts, employ 
non-standard Romanization, and embed culture-specific 
slang that translation systems cannot reliably 
interpret~\cite{santos2023capivara}. This brittleness 
is especially acute for Dravidian languages such as 
Tamil and Malayalam, where Tanglish and Manglish 
conventions produce out-of-vocabulary tokens that 
push multilingual encoders outside their pretraining 
distribution, degrading the textual embeddings that 
downstream classifiers depend on.

\section{Methodology}
\label{sec:method}

\begin{figure*}[htbp]
  \centering
  \includegraphics[width=0.8\linewidth, height=8cm, keepaspectratio]{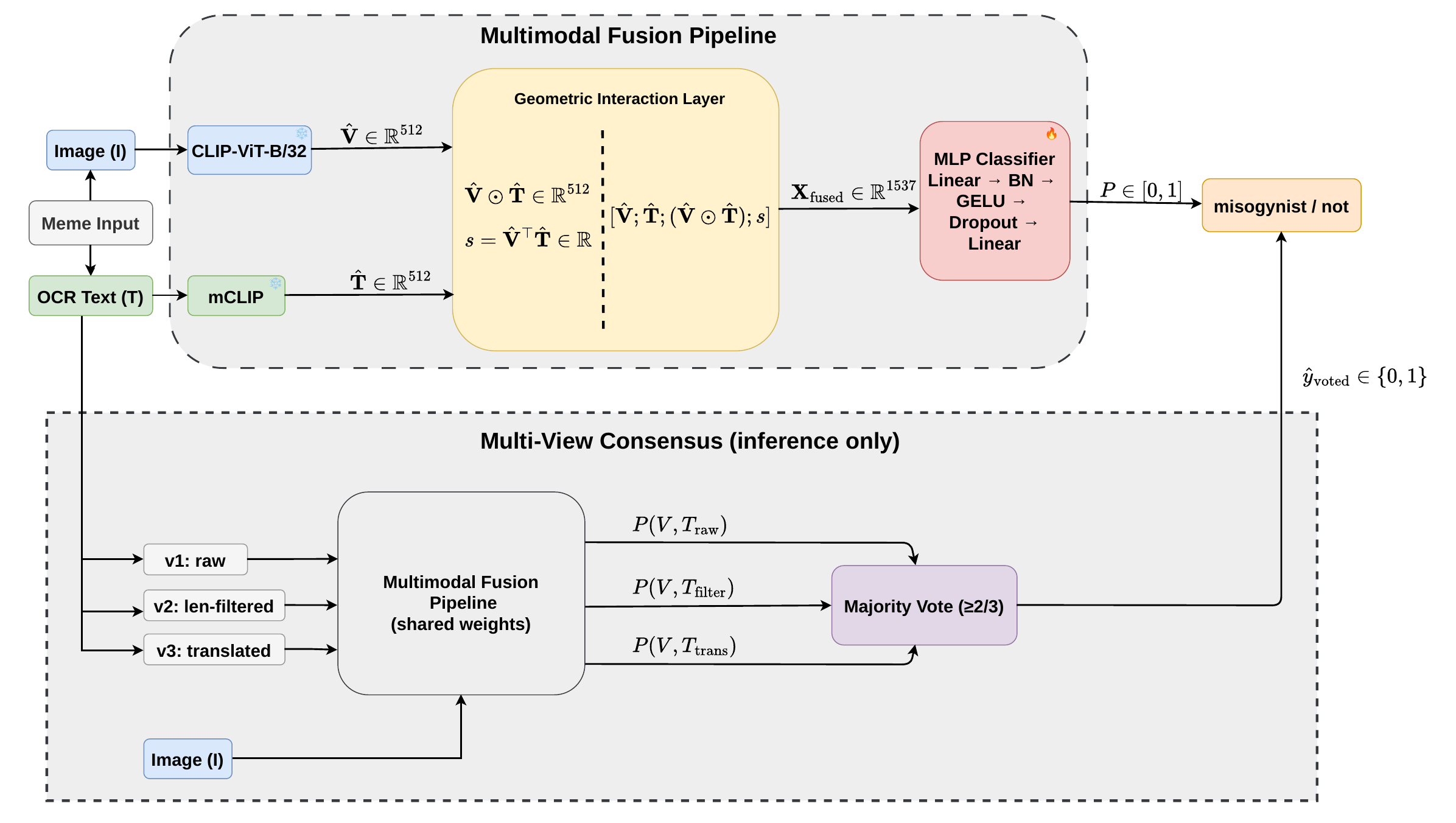}
  \caption{GeoMVC system architecture. Top row: training and single-view
  inference. Bottom row: Multi-View Consensus path (inference only).}
  \label{fig:arch}
\end{figure*}

The GeoMVC framework consists of three components: a dual-encoder
feature extractor, a Geometric Interaction Layer, and a Multi-View
Consensus inference strategy. Figure~\ref{fig:arch} illustrates
the full pipeline.

\subsection{Feature Extraction}
\label{sec:features}

Given a meme comprising an image $I$ and OCR-transcribed text $T$,
we extract embeddings using two frozen encoders. CLIP-ViT-B/32~\cite{radford2021learning}
produces a visual embedding $\mathbf{V} \in \mathbb{R}^{512}$.
Multilingual CLIP (mCLIP)~\cite{carlsson2022cross} produces a
textual embedding $\mathbf{T} \in \mathbb{R}^{512}$, after
first cleaning $T$ by stripping URLs, handles, and hashtags.
Only the Geometric Interaction Layer and classifier are updated
during training; both encoders remain frozen.

\subsection{Geometric Interaction Layer}
\label{sec:interaction}

Both embeddings are L2-normalized to
$\hat{\mathbf{V}}$ and $\hat{\mathbf{T}}$.
Four components are constructed: the visual prior $\hat{\mathbf{V}}$,
the textual prior $\hat{\mathbf{T}}$, their Hadamard product
$\hat{\mathbf{V}} \odot \hat{\mathbf{T}} \in \mathbb{R}^{512}$
(amplifying dimensions of cross-modal agreement or disparity),
and the cosine scalar
$s = \hat{\mathbf{V}}^{\top}\hat{\mathbf{T}} \in \mathbb{R}$,
which follows directly from the normalization. These are
concatenated as:
\vspace{-2mm}
\begin{equation}
  \mathbf{X}_{\text{fused}} =
  \bigl[\hat{\mathbf{V}};\,\hat{\mathbf{T}};\,
  (\hat{\mathbf{V}} \odot \hat{\mathbf{T}});\,s\bigr]
  \in \mathbb{R}^{1537}
\end{equation}

$\mathbf{X}_{\text{fused}}$ is passed through a lightweight MLP:
\begin{equation}
  \hat{y} = \sigma\!\left(
    W_2\,\mathrm{Dropout}\!\left(
      \mathrm{GELU}\!\left(
        \mathrm{BN}(W_1\,\mathbf{X}_{\text{fused}} + b_1)
      \right)
    \right) + b_2
  \right)
\end{equation}

where $W_1 \in \mathbb{R}^{256 \times 1537}$,
$W_2 \in \mathbb{R}^{1 \times 256}$, BN denotes Batch Normalization,
and $\sigma$ is the sigmoid function.

\subsection{Training Objective}
\label{sec:training}

We train partition-specific specialists (one for Malayalam/India,
one for Chinese) using Binary Cross-Entropy with a positive class
weight to address label imbalance:
\vspace{-2mm}
\begin{equation}
  \mathcal{L} = -\frac{1}{N}\sum_{i=1}^{N}
  \Bigl[
    w^{+} \cdot y_i \log \hat{y}_i +
    (1-y_i)\log(1-\hat{y}_i)
  \Bigr]
\end{equation}

where $w^{+} = 2.0$ upweights the misogynist class. We use AdamW
($\eta = 10^{-4}$) with mixed-precision training via PyTorch AMP.
The model is trained for 50 epochs; the checkpoint with the highest
Macro F1 on the development set is retained for inference.

\subsection{Multi-View Consensus}
\label{sec:consensus}

OCR outputs in low-resource scripts are noisy and inconsistent.
To improve inference robustness, each test sample's text is
branched into three parallel views before prediction:

\begin{itemize}
  \item $V_{\text{raw}}$: the cleaned OCR transcription as-is.
  \item $V_{\text{filter}}$: a length-filtered variant retaining
        only tokens with more than two characters, reducing the
        influence of abbreviations and single-character noise
        common in Romanized Dravidian scripts.
  \item $V_{\text{trans}}$: an English back-translation obtained
        via Google Translate, intended to leverage the stronger
        high-resource alignment of mCLIP.
\end{itemize}
 Each view is independently scored, and the final
label is determined by majority vote:
\vspace{-2.5mm}
\begin{equation}
  \hat{y}_{\text{voted}} =
  \mathbf{1}\!\left[\sum_{k} p_k \geq 2\right]
\end{equation}
This ensemble-like mechanism acts as a regularizer at deployment
time, preventing a single corrupted OCR view from dominating the
decision.
\section{Experimental Setup}
\label{sec:exp}

\subsection{Dataset and Task Definition}

\begin{table}[htbp]
  \caption{Label distribution across partitions and splits}
  \label{tab:distribution}
  \resizebox{\columnwidth}{!}{%
  \begin{tabular}{llrrrr}
    \toprule
    Partition & Split & Total & Misogynist & Not-misogynist & Ratio \\
    \midrule
    \multirow{2}{*}{Malayalam} 
      & Train & 640  & 256 (40.0\%) & 384 (60.0\%) & 1.50x \\
      & Dev   & 160  & 64 (40.0\%)  & 96 (60.0\%)  & 1.50x \\
    \midrule
    \multirow{2}{*}{Tamil}
      & Train & 1137 & 274 (24.1\%) & 863 (75.9\%) & 3.15x \\
      & Dev   & 284  & 75 (26.4\%)  & 209 (73.6\%) & 2.79x \\
    \midrule
    \multirow{2}{*}{Chinese}
      & Train & 1190 & 349 (29.3\%) & 841 (70.7\%) & 2.41x \\
      & Dev   & 170  & 47 (27.6\%)  & 123 (72.4\%) & 2.62x \\
    \bottomrule
  \end{tabular}%
  }
  \vspace{-4mm} 
\end{table}

GeoMVC was evaluated on the CC-MMD dataset~\cite{ponnusamy2024laughter,
chakravarthi2025overview, fersini2022semeval} provided for the ICMI 
2026 Grand Challenge, focusing on Task~A (binary misogyny 
classification) under the Original Culture label scheme. CC-MMD 
is a multilingual, multimodal benchmark integrating memes across 
three culture-specific partitions: Indian context (Tamil and 
Malayalam, sourced from MDMD~\cite{ponnusamy2024laughter}), 
Chinese context (sourced from CMMD~\cite{chakravarthi2025overview}), 
and Western context (English, sourced from 
MAMI~\cite{fersini2022semeval}). 
Table~\ref{tab:distribution} reveals substantial class imbalance 
across partitions, with Tamil exhibiting the most severe ratio 
(3.15x in train) and Malayalam the most balanced (1.50x). 
The train and dev distributions are consistent within each 
partition, indicating no significant sampling bias between splits.

\subsection{Implementation Details}
\label{sec:impl}

The system was implemented in PyTorch and trained on a single 
NVIDIA Tesla T4 GPU via Google Colab. Visual and textual features
were extracted once and cached on disk to avoid redundant computation
across training runs. Table~\ref{tab:impl} summarises the key
hyperparameters.

\begin{table}[htbp]
  \caption{Implementation hyperparameters}
  \label{tab:impl}
  \begin{tabular}{ll}
    \toprule
    Hyperparameter        & Value \\
    \midrule
    Visual encoder        & CLIP-ViT-B/32 (frozen) \\
    Textual encoder       & mCLIP ViT-B/32 (frozen) \\
    Fused dimension       & 1537 \\
    MLP hidden units      & 256 \\
    Dropout               & 0.4 \\
    Pos. class weight     & 2.0 \\
    Optimizer / LR        & AdamW / $1\times10^{-4}$ \\
    Batch size / Epochs   & 64 / 50 \\
    Selection criterion   & Best Macro F1 (dev) \\
    \bottomrule
  \end{tabular}
\end{table}

\section{Results}

\label{sec:results}

\subsection{Official Leaderboard Performance}
\label{sec:leaderboard}

Table~\ref{tab:results} reports the GeoMVC system's performance
on the hidden test set for Task A across all three partitions.

\begin{table}[h]
  \caption{Official leaderboard results --- GeoMVC system on
  Original Culture (Task A)}
  \label{tab:results}
  \begin{tabular}{lccc}
    \toprule
    Partition & Accuracy & Macro F1 & Rank \\
    \midrule
    Malayalam & 0.895 & 0.892 & \textbf{2} \\
    Chinese   & 0.912 & 0.895 & \textbf{3} \\
    Tamil     & 0.640 & 0.521 & 5 \\
    \bottomrule
  \end{tabular}
\end{table}

The system achieved podium finishes in both the Malayalam and
Chinese partitions. The Malayalam F1 of 0.892 placed GeoMVC
within a marginal gap of the first-ranked system, offering evidence that
the Geometric Interaction Layer's capacity to model implicit
cross-modal semantic clash effectively. The higher Chinese accuracy (0.912)
alongside a slightly lower F1 (0.895) suggests a degree of
class imbalance in the Chinese test partition.
Table~\ref{tab:ablation} shows the contribution of each 
component on the Malayalam development set. While adding 
the Hadamard product alone yields marginal change, combining 
it with the cosine similarity scalar provides a clear 
improvement, demonstrating the value of the full Geometric 
Interaction Layer.

\begin{table}[h]
  \caption{Ablation study on Malayalam dev set with Multi-View
  Consensus (Macro F1). GeoMVC achieves the highest F1,
  with the cosine scalar providing the most meaningful gain,
  particularly under noisy OCR conditions.}
  \label{tab:ablation}
  \begin{tabular}{lc}
    \toprule
    Configuration & Macro F1 \\
    \midrule
    Concatenation only           & 0.890 \\
    $+$ Hadamard product         & 0.888 \\
    $+$ Cosine scalar (GeoMVC)   & \textbf{0.903} \\
    \bottomrule
  \end{tabular}
\end{table}


\section{Error Analysis}

\subsection{Class-Level Breakdown}
\label{sec:classbreak}

Table~\ref{tab:error} presents the confusion matrices and
multi-view disagreement statistics across all three dev sets.
Error profiles differ meaningfully by partition. Malayalam
exhibits more false negatives than false positives (FN=10,
FP=8), consistent with the model being conservative in flagging
misogyny. Tamil shows the opposite pattern (FN=20, FP=29),
where the severe class imbalance (3.15x ratio in training)
causes the model to over-predict misogyny on ambiguous inputs.
Chinese achieves the cleanest separation with only 8 total
errors across 170 samples.

\begin{table}[h]
  \caption{Cross-partition error analysis on dev sets}
  \label{tab:error}
  \begin{tabular}{lcccccc}
    \toprule
    Partition & TP & TN & FP & FN & 2-1 splits & Wrong splits \\
    \midrule
    Malayalam & 54 & 88  & 8  & 10 & 12 (7.5\%) & 8 (66.7\%) \\
    Tamil     & 55 & 180 & 29 & 20 & 17 (6.0\%) & 9 (52.9\%) \\
    Chinese   & 44 & 118 & 5  & 3  & 13 (7.6\%) & 4 (30.8\%) \\
    \bottomrule
  \end{tabular}
\end{table}

\subsection{Cross-Modal Cosine Similarity}
\label{sec:cosine}

We computed the cross-modal cosine similarity $s$ for every
development sample across outcome groups. In isolation, cosine
similarity showed negligible discriminative power---the difference
between group means was under 0.03 in every partition. This confirms
that global semantic alignment alone cannot distinguish implicit
misogyny. However, as demonstrated in Table~\ref{tab:ablation},
including $s$ yields a clear F1 improvement. We hypothesize that
while weak as a standalone metric, the downstream MLP effectively
utilizes the cosine scalar in combination with the Hadamard product,
acting as a global contextual scaling factor for the element-wise clash.

\subsection{Multi-View Disagreement}
\label{sec:disagreement}

The 2-1 split rate is consistent across partitions (6.0--7.6\%),
indicating similar levels of view disagreement regardless of
language. However, the proportion of wrong decisions among
contested cases drops sharply from 66.7\% in Malayalam and
52.9\% in Tamil to just 30.8\% in Chinese. When
$V_{\text{trans}}$ was the dissenting vote, the majority was
correct in only 4/11 Malayalam and 6/13 Tamil cases, but 8/12
Chinese cases. This inversion suggests Google Translate
preserves semantic fidelity for Chinese far better than for
code-mixed Dravidian scripts, where transliteration noise
degrades all three views simultaneously.

\subsection{Tamil Partition: Primary Failure Modes}
\label{sec:tamil}

The Tamil test F1 of 0.521 reflects a field-wide challenge
rather than a system-specific failure. Two primary causes
emerge from dev set analysis.

\textbf{Class imbalance.} With a 3.15x train imbalance,
the model over-generates false positives on culturally
ambiguous Tamil content despite the $w^{+}=2.0$ correction.
The 29 FP versus 20 FN on the dev set confirms this bias.

\textbf{Extreme code-mixing.} Tamil memes use heavy Tanglish
--- Roman-script Tamil interleaved with English slang. The
mCLIP encoder inconsistently maps out-of-vocabulary
transliterations to correct semantic regions, producing
noisy $\mathbf{T}$ embeddings. The translated view
$V_{\text{trans}}$ frequently loses sarcastic undertone,
causing flawed majority votes in exactly the cases where
consensus is most needed.

\section{Discussion}
\label{sec:discussion}

The results demonstrate that explicit geometric interaction provides a meaningful signal beyond simple late fusion. The Hadamard product effectively amplifies cross-modal agreement while suppressing neutral dimensions, whereas global cosine similarity adds negligible value. Multi-View Consensus improves robustness against noisy OCR and transliteration in Malayalam and Chinese partitions. However, it cannot fully compensate for systematic encoder limitations under extreme Tanglish code-mixing, as observed in the Tamil partition. These findings highlight the persistent challenges of multilingual VLMs in heavily code-mixed, low-resource settings.

\section{Conclusion}
\label{sec:conclusion}

This paper presents GeoMVC, a practical system for cross-cultural 
multimodal misogyny detection in memes. By integrating a Geometric 
Interaction Layer with Hadamard products and a Multi-View Consensus 
strategy, our approach achieved strong results on the CC-MMD 2026 
challenge --- securing Rank~2 in Malayalam (Macro F1: 0.892) and 
Rank~3 in Chinese (Macro F1: 0.895). Error analysis highlights 
that element-wise geometric interactions effectively capture 
implicit semantic clash, while global cosine similarity provides 
limited discriminative value. Nevertheless, extreme code-mixing in Tamil remains challenging
when relying on frozen multilingual VLMs. Future work will explore
parameter-efficient fine-tuning (e.g., LoRA) on the text encoders
and transliteration-aware adapters to address these low-resource
limitations directly during training.

\begin{acks}
We thank the organizers of the CC-MMD Grand Challenge for 
providing the benchmark datasets and supporting this research.
\end{acks}


\begin{thebibliography}{99}

\bibitem{kiela2020hateful}
Douwe Kiela, Hamed Firooz, Aravind Mohan, Vedanuj Goswami,
Amanpreet Singh, Pratik Ringshia, and Davide Testuggine. 2020.
The Hateful Memes Challenge: Detecting Hate Speech in Multimodal
Memes. In \emph{Advances in Neural Information Processing Systems
(NeurIPS)}, Vol.~33, 2611--2624.

\bibitem{radford2021learning}
Alec Radford, Jong Wook Kim, Chris Hallacy, Aditya Ramesh,
Gabriel Goh, Sandhini Agarwal, Girish Sastry, Amanda Askell,
Pamela Mishkin, Jack Clark, et al. 2021. Learning Transferable
Visual Models From Natural Language Supervision. In
\emph{International Conference on Machine Learning (ICML)},
8748--8763.

\bibitem{carlsson2022cross}
Fredrik Carlsson, Philipp Eisen, Faton Rekathati, and Magnus
Sahlgren. 2022. Cross-lingual and Multilingual CLIP. In
\emph{Proceedings of the Language Resources and Evaluation
Conference (LREC)}, 6848--6854.

\bibitem{Ben-Younes2017MUTAN}
Hedi Ben-Younes, R\'{e}mi Cadène, Matthieu Cord, and Nicolas
Thom\'{e}. 2017. MUTAN: Multimodal Tucker Fusion for Visual
Question Answering. In \emph{Proceedings of the IEEE International
Conference on Computer Vision (ICCV)}, 2612--2620.

\bibitem{Ben-Younes2019BLOCK:}
Hedi Ben-Younes, R\'{e}mi Cadène, Nicolas Thom\'{e}, and Matthieu
Cord. 2019. BLOCK: Bilinear Superdiagonal Fusion for Visual Question
Answering and Visual Relationship Detection. In \emph{Proceedings
of the AAAI Conference on Artificial Intelligence}, Vol.~33,
8102--8109.

\bibitem{Cadène2019MUREL:}
R\'{e}mi Cadène, Hedi Ben-Younes, Matthieu Cord, and Nicolas
Thom\'{e}. 2019. MUREL: Multimodal Relational Reasoning for
Visual Question Answering. In \emph{Proceedings of the IEEE
Conference on Computer Vision and Pattern Recognition (CVPR)},
1989--1998.

\bibitem{chen2023mclip}
Guanhua Chen, Lu Hou, Yun Chen, Wenliang Dai, Lifeng Shang,
Xin Jiang, Qun Liu, Jia-Yu Pan, and Wenping Wang. 2023.
mCLIP: Multilingual CLIP via Cross-lingual Transfer. In
\emph{Proceedings of the 61st Annual Meeting of the Association
for Computational Linguistics (ACL)}, 13028--13043.

\bibitem{santos2023capivara}
Guilherme Santos, Diego Moreira, Alan Ferreira, Jhessica Silva,
Luiz Pereira, Pedro Bueno, Thiago Sousa, Hasan Maia, N\'{a}dia
Silva, Esther Colombini, H\'{e}lio Pedrini, and Sandra Avila.
2023. CAPIVARA: Cost-Efficient Approach for Improving Multilingual
CLIP Performance on Low-Resource Languages. \emph{arXiv preprint
arXiv:2310.13683}.

\bibitem{zhong2020classification}
Xiayu Zhong. 2020. Classification of Multimodal Hate Speech ---
The Winning Solution of Hateful Memes Challenge. \emph{arXiv
preprint arXiv:2012.01002}.

\bibitem{lippe2020multimodal}
Phillip Lippe, Nithin Holla, Shantanu Chandra, Santhosh
Rajamanickam, Georgios Antoniou, Ekaterina Shutova, and Helen
Yannakoudakis. 2020. A Multimodal Framework for the Detection
of Hateful Memes. \emph{arXiv preprint arXiv:2012.12871}.

\bibitem{velioglu2020detecting}
Riza Velioglu and Jewgeni Rose. 2020. Detecting Hate Speech
in Memes Using Multimodal Deep Learning Approaches:
Prize-winning Solution to Hateful Memes Challenge.
\emph{arXiv preprint arXiv:2012.12975}.
\bibitem{fersini2022semeval}
Elisabetta Fersini, Francesca Gasparini, Giulia Rizzi, Aurora Saibene,
Berta Chulvi, Paolo Rosso, Alyssa Lees, and Jeffrey Sorensen. 2022.
SemEval-2022 Task 5: Multimedia Automatic Misogyny Identification.
In \emph{Proceedings of the 16th International Workshop on Semantic
Evaluation (SemEval-2022)}, 533--549.

\bibitem{ponnusamy2024laughter}
Rahul Ponnusamy, Kathiravan Pannerselvam, Saranya Rajiakodi,
Prasanna Kumar Kumaresan, Sajeetha Thavareesan, Bhuvaneswari
Sivagnanam, Anshid K.A, Susminu S Kumar, Paul Buitelaar, and
Bharathi Raja Chakravarthi. 2024. From Laughter to Inequality:
Annotated Dataset for Misogyny Detection in Tamil and Malayalam
Memes. In \emph{Proceedings of the 2024 Joint International
Conference on Computational Linguistics, Language Resources and
Evaluation (LREC-COLING 2024)}, 7480--7488.

\bibitem{chakravarthi2025overview}
Bharathi Raja Chakravarthi, Rahul Ponnusamy, Ping Du, Xiaojian
Zhuang, Saranya Rajiakodi, Paul Buitelaar, Premjith B,
Bhuvaneswari Sivagnanam, Anshid Kizhakkeparambil, and Lavanya
S.K. 2025. An Overview of the Misogyny Meme Detection Shared
Task for Chinese Social Media. In \emph{Proceedings of the 5th
Conference on Language, Data and Knowledge: Fifth Workshop on
Language Technology for Equality, Diversity, Inclusion}, 200--208.

\end{thebibliography}
\end{document}